# Web-Based Implementation of Travelling Salesperson Problem Using Genetic Algorithm

A Comparative Study of Python, PHP, and Ruby

Aryo Pinandito[1], Novanto Yudistira[2], Fajar Pradana[3]
Program Teknologi Informasi dan Ilmu Komputer[1]
Universitas Brawijaya
Malang, Indonesia
aryo@ub.ac.id[1], yudistira@ub.ac.id[2], fajar.p@ub.ac.id[3]

*Abstract*—The world is connected through the Internet. As the abundance of Internet users connected into the Web and the popularity of cloud computing research, the need of Artificial Intelligence (AI) is demanding. In this research, Genetic Algorithm (GA) as AI optimization method through natural selection and genetic evolution is utilized. There are many applications of GA such as web mining, load balancing, routing, and scheduling or web service selection. Hence, it is a challenging task to discover whether the code mainly server side and web based language technology affects the performance of GA. Travelling Salesperson Problem (TSP) as Non Polynomial-hard (NP-hard) problem is provided to be a problem domain to be solved by GA. While many scientists prefer Python in GA implementation, another popular high-level interpreter programming language such as PHP (PHP Hypertext Preprocessor) and Ruby were benchmarked. Line of codes, file sizes, and performances based on GA implementation and runtime were found varies among these programming languages. Based on the result, the use of Ruby in GA implementation is recommended.

*Keywords—TSP; Genetic Algorithm; web-programming language*

## I. Introduction

AI refers to intelligent behaviour unexceptionally in web-based application. A combination of the web application and AI has been becoming the future trends of web-based application [1]. Moreover, the trend of cloud computing has already risen [2]. There are many web-based applications use GA for various purposes. GA as a well known algorithm that use heuristic approach to gather fully optimized solutions, has been used widely in various web applications such as search engine and web mining application [3]. GA has become the effective algorithm in terms of pattern recognition. Recently, it is found that new trend like social graph technology with optimization is promising [4].

From the scientific point of view, data processing and analysis scripts often time consuming and require many hours to be computed on a computer device, so the iteration process along with its debugging process will be longer [5]. Moreover, scientists have a different focus on his work compared to professional programmers. They are keen on methodology rather than the tools they are utilizing.

Faster completion of programming task is surely dreamed by many scientists or even beginner programmers who have a consideration of being effective. Naturally, they will choose that kind of Programming Language (PL). It is also based on a psychological review as narrated in [5]. To catch up with the rapidly progressive research in AI, speed and simplicity become necessary in AI programming. Therefore it is needed to give scientists PLs that could quickly iterate while preserve the tidiness and simplicity so that it can be easily used.

Even though compiled PL is faster at run time than interpreter PL, it is not as simple as recent emerging PL such as Python [6], PHP [7] and Ruby [8]. The advantages of using Python basically lie in its ease of use, interpreted and object oriented programming language that can bridge many scientists need without loosing the sense of object oriented style. However, the effectiveness of PL can be measured by how many lines of code should be written or how much syntaxes should be initiated to implement the same GA. Another drawback of using compiled PL is by looking at the denial of service type of attacks [9].

## II. Computational

There are many researches use GA for benchmarking purposes [10]. This paper benchmarks interpreter PL in supporting AI. Problem domain to be solved by GA in this case is Travelling Salesperson Problem (TSP) in which has become a benchmark for several heuristics in GA performance test [11]. The use of TSP varies based on its domain problems. TSP is an NP-hard problem which solution is optimized by GA based on natural selections and genetic evolutions to solve NP complete domain problems.

At TSP, it is the idea of finding a route of a given number of cities by visiting each city exactly once and return to the starting city where the length of the route is minimized. A path that visits every city and returns to the starting city creating a closed circuit is called a route. The simplest and direct method to solve TSP of any number of cities is by enumerating every possible route, calculating every route length, and choosing one route with the shortest length. It is possible that every city may become the starting point of the route and one route may have the same length regardless of the direction of the route taken. Logically, the problem can be set up by using integer $n>0$ and the distance between every pair of n cities and represented by an $n \times n$ matrix [11]. Each possible route can be represented as a permutation of n, where n is the number of cities, thus the

This research supported by Program Teknologi Informasi dan Ilmu Komputer (PTIIK), Universitas Brawijaya.

number of possible routes is a factorial of n (n!). It turns out to be more computationally burdened and difficult to enumerate and find the length of every route as n may become larger in polynomial complexity. Hence, TSP needs an algorithm that able to find a route that produces the minimum length without having to enumerate all possible routes from cities given.

Genetic Algorithm (GA) is basically a search algorithm that is used to find solutions of evolution in such search space to solve problems such as TSP. To this end, a solution is so called an individual while a set of solutions is normally called as population. Individuals are evolving through many generations in populations. The important thing that must be initiated is the genetic representation of the solution domain and the fitness function in order to evaluate the quality of such individuals [12]. Moreover, selection is a stochastic function. It selects individuals based on the fitness value of a generated population, and yields so called the best-selected individual.

Even though GA that utilizes crossover and mutation steps does not define the end of the iteration, every generation has the probability to obtain better individuals than its ancestors in terms of their fitness cost. It has been a state-of-the-art that solving the TSP problem using GA is optimal but it depends on crossover and mutation method used in this research.

*A. Crossover*

Crossover is the step to produce various chromosomes (individuals), which is chosen from the chromosomes in previous generation. In this research, selecting two best chromosomes that have the best fitness cost to crossover does the crossover steps. Then, randomly picking a gene or point from both chromosomes, and finally pick the rest of remaining genes randomly until all unique genes has been picked up. A new population later will be generated from previously selected chromosomes. Therefore, recombination process of two parent individuals will generate new chromosomes. An example in producing an individual from two individuals as parents, which represented using "a" and "b", by doing crossover, is explained below:

a = ABCDEFGH
b = EFGHABCD

Each gene of a new individual was taken from both parent genes. The crossover process is done until it generates a new individual "c" like the following example:

c = AEBCGFDH

*B. Mutation*

Mutation is an extension of the crossover process that executed by such a probability rate. It is used to avoid a local optimum. If the mating process only depends on crossover, it probably yields to a local optimum because the chance to approach the fitness value is relatively high. Mutation process is given to cover the problem in such a way to reach global optimum solution.

The mutation method used in this research was done by choosing a gene in such an index and switch the pointed gene to be the first index along with the rest of the genes that follows in the sequence. For example, if an offspring "c" produced by the crossover process is "ABCDEFGH", and the pointed gene for mutation is E, then the resulting mutation offspring is "EFGHABCD".

III. INTERPRETED LANGUAGES

PLs are related to programmable and dynamic environment in which components were bound together at a high level [13]. The emergence of compiled languages such as Java or C has led to be World Wide Web (WWW) transformation [9]. However, these PL leave drawbacks such as parsed codes produced by compilers are stored in scattered files. Hence, in a compiled environment, some of those files must be included such as the instruction codes, parsed codes, header files, and the executable or linked codes altogether. Nonetheless, the significant distinction of compilers and interpreters is that compilers parse and execute in different actions sequentially. Though its speed and standalone executable existence of compiled PL, there is an increasing complexity in its process. Nevertheless, many recent compilers are able to compile and execute the code directly in memory, giving such an interpreted language (IL) fashion [14].

The use of interpreted PL is beneficial since in an interpreted environment every instruction of PL is executed right away after being parsed. This has eased the programmers since the result can be presented quickly. Moreover, the interpreted code can be run without compilers and linkers to produce executable codes. However, there are disadvantages related to IL such as its poor performance, no executable program, and interpreter dependence compared to a compiled PL [14]. After all, ILs such as PHP, Python, and Ruby were merely made by hobbyists without any long-term research goal such as Lisp but they are widely applicable.

*A. Interpreter Support for GA*

There are many GA implementations built into Python such as pyGA. The exact result of effectiveness between Python, PHP, and Ruby in GA implementation is still a domain of interest to be explored. GA, despite of its pros and cons, is quite easy to be implemented, but yields into a very slow process in its usage to solve such problems.

IV. METHODOLOGY

TSP based GA in this research is implemented into three commonly used server-side object oriented programming language and without any frameworks. They are Python, PHP, and Ruby. All of GA codes that represent each PL are written based on the given pseudo code. They were implemented using the same variable names, methods, and initialization logic. During implementation, they were tested using one data source and the same parameter values for all PL.

GA codes of each PL are implemented as close as possible. If any parts of pseudo code are implemented on one PL using multiple methods, functions, or variables, then the other PL had to be implemented in the same manner. Keeping the code to be as close as possible supposedly yields to objective measurement results.

One of many important functions used in GA is a random number generator. In GA, random number generator functions utilized to generate a random number in order to compensate the probability of doing a crossover, copy, or mutate the parents. The random number generator is implemented using a

Pseudo-Random Number Generator (PRNG) function. PRNG is a random number generator function in which when it was given the same seed number, it will always return the same random value. This technique is used to ensure that all programs will go through the same method and loop inside the program when run under the same circumstances, thus giving the same result. The PRNG function used in this research is implemented using a separate script that is run by a system call by each of implemented PL.

In performance measurement of implemented PL, a small modification was added to the scripts by adding current time or micro time function at several points in code while taking account into current generation best fitness cost value. Before carrying out any measurement values, testing units were made using a specified value and data to ensure all implemented scripts are using the same seed number and random values, therefore all implemented PL will mate the equal parents and generating the same candidate population in every generation. They should return the same best individual in the same generation at the end of script execution.

The scripts were run and measured on a MacBook Pro computer running on Mac OS X 10.8.2, 2.5 GHz Intel Core i5 processor, 8 GB 1600 MHz DDR3 memory, and Intel HD 4000 graphics with 512 MB shared-memory. The version of Python, Ruby, and PHP interpreter used in measurement processes are 2.7.3, 1.9.3, and 5.3.15 respectively.

The main objective of this paper is to automatically infer more precise bounds on execution times and best fitness that depends on input data sizes of the three different PL. Thus, a recommendation to a widely used server-side PL in GA implementation is given.

*A. Genetic Algorithm Pseudo Code for TSP*

TABLE I shows how GA can solve TSP. It will be written in Python, PHP, and Ruby. As of the initial population, individuals are generated from a Comma Separated Value (CSV) file that contains names of cities along with its x and y coordinates.

TABLE I. GENETIC ALGORITHM PSEUDO CODE

| GA Pseudo Code |
|---|
| **class** City<br>  **function** initialize (name, x, y)<br>    Initialize name of city along with its coordinate in x and y<br>**class** Individual<br>  **function** initialize (route)<br>    Initialization of an individual which consists of cities or nodes<br>  **function** makeChromosome (file)<br>    Generate an individual chromosomes from a file<br>  **function** evaluate<br>    Calculate the length of given population (a set of cities) using<br>    Euclid's distance formula<br>  **function** crossover (other)<br>    Do a recombination to return a new offspring from given spouse<br>**class** Environment<br>  **function** initialize<br>    Initialization to population data, population size, maximum number<br>    of generations, crossover rate, mutation rate, and optimum number<br>    of generation related to the best fitness cost convergence value<br>  **function** makePopulation<br>    Create an initial population consists of two parents, from file<br>    (as parent A), other parent is generated from shuffled parent A<br>  **function** evaluate<br>    Sorts all individual based on the its calculated length value, starting<br>    from the least length (best) until the largest individual in current<br>    population.<br>  **Do** getBestIndividual<br>  **function** getBestIndividual<br>    **if** the best length obtained so far is less than the best individual in<br>    population obtained from crossover<br>    **then** the best fitness cost remains unchanged<br>    **else** the best fitness cost obtained is updated from the first individual<br>    order of population in the current crossover's offspring population.<br>  **function** run<br>    **while** the goal is not achieved<br>      **Do** a generation step<br>    Increase generation number by 1<br>  **function** goal<br>    **if** current generation reached maximum generation or current<br>    generation reached optimum (when current generation minus best<br>    individual generation is larger than optimum number given)<br>    **then** the goal has been achieved otherwise continue iteration.<br>  **function** step<br>    **Do** a crossover<br>    Evaluate fitness costs of all individual in the current population<br>  **function** crossover<br>    Initialize candidate population<br>    Select two best parent candidates (parent A and parent B) from<br>    current population.<br>    **While** candidate population is still less than given size, **do**<br>    randomise crossover rate<br>    **if** in crossover rate<br>    **then** offspring = crossover parent A with parent B<br>    **else** offspring = copy parent A<br>    randomise mutation rate<br>    **if** in mutation rate<br>    **then** offspring = mutate current offspring<br>    evaluate offspring's fitness cost<br>    **if** offspring not exists in candidate population<br>    **then** add offspring (individual) into candidate population<br>    Candidate population become the new population<br>  **function** mutate (individual)<br>    Mutating individual in the manner of switching half of given<br>    individual orderly<br>**Run** GA environment with parameters: CSV file containing population data, number of population, number of maximum generation, crossover rate, mutation rate, and optimum best generation. |

*B. PL Implementation of GA Pseudo Code*

GA pseudo code used to solve TSP in this research was implemented in Python, PHP, and Ruby. Every method and variables are unit-tested, ensuring they have the same results and values across all implementations of pseudo-code given. The codes are utilizing the same population data.

It is impossible to implement methods and variables in the exact ways due to the different natures of programming style of each PL. However, all implemented codes of methods and variables across all PL are ensured to behave the same and having the same values under the same GA environment variables and population data.

*C. Random Number Generation*

As previously mentioned, the random number generator such as PRNG is used in GA are implemented using a separate code. It is called by a system call function from the implemented PL. therefore all implemented scripts will have the same random number performance and value under the same circumstances. The PRNG code is implemented in Ruby because of its known good performance [15] and it is shown in TABLE II.

Native random number generator functions across different PL behave differently. Separation of PRNG code from main script is required to ensure all PL are using the same random number during running time and therefore the execution time measurement would be objective. In real world GA implementation, the use of native random function is recommended. This research proved that the use of system call causes bottleneck in program execution.

TABLE II. PRNG IMPLEMENTATION IN RUBY

| PRNG Code Implementation in Ruby |
|---|
| ```
if(ARGV.length == 2)
  seed = ARGV[0].to_i
  max = ARGV[1].to_i
  srand(seed)
  if(ARGV[1].to_i == 1)
    puts rand
  else
    puts rand(ARGV[1].to_i)
  end
else
  puts 0
end.
``` |

*D. Seed Number Generation*

The seed number, which is used to generate random values, is incremented by one (+1 from the previous seed value) before calling the PRNG script. The seed itself implemented as a global variable, therefore its value will always be available before it is fed to the PRNG and its value will be maintained during runtime.

*E. Data*

In this experiment, genome data for an individual that used in unit tests and measurements are shown on TABLE III. They are stored in a CSV file as a plain text. Values stored on each line in a CSV file consist of the name of cities along with their coordinates in X and Y-axis. The file is read during initialization of GA environment and its values are used as population's initial genome data.

TABLE III. INITIAL POPULATIONS OF GNOME DATA

| No | City Name | X | Y |
|---|---|---|---|
| 1 | Balikpapan | 5 | 4.2 |
| 2 | Malang | 9.2 | 1.1 |
| 3 | Jayapura | 16.4 | 8.2 |
| 4 | Manado | 10.3 | 9.7 |
| 5 | Bandung | 6 | 1.3 |
| 6 | Banjarmasin | 5 | 3.1 |
| 7 | Pontianak | 5.3 | 5.2 |
| 8 | Jakarta | 5.3 | 2.3 |
| 9 | Medan | 2.1 | 10.6 |
| 10 | Makassar | 10 | 4 |

*F. PL Execution Time Measurement*

Workarounds have to be used in order to do a portable time measurement where high-resolution timing is difficult or impossible to achieve. Synthetic benchmark approach were followed, which on purpose repeatedly execute the instructions under estimation for a large enough time, and later averaging the total execution time by the number of times it is being run. Generally, it is not possible to run a single instruction repeatedly within the abstract machine, since the resulting sequence would not be legal and may "break" the abstract machine, run out of memory, etc. Therefore, more complex sequences of instructions must be constructed and be repeated instead.

As of previous measurement research conducted in the case of 0/1 Knapsack problem [16] were measured by using each PL's native timing function. The execution time measurements in our TSP problem were also doing so.

In getting program execution time, the difference between the start time and end time of script execution is calculated. These measurements are done several times under the same environment circumstances on all PL. Each measurement is done with different number of population data from CSV file.

Program execution times are measured using five, six, seven, eight, nine, and ten cities in consecutive ways. GA environmental parameters were set to 100 maximum generations, 5 populations within each generation, with a 90% crossover rate, 1% mutation rate, and 20 generations limit of best fitness cost as a GA convergence termination limit. Each GA environment is measured 10 times, and then the average values and their standard deviations are estimated.

V. RESULT AND ANALYSIS

Implementation of GA pseudo code provided is resulting in Python, PHP, and Ruby codes. Python, PHP, and Ruby codes were implemented in 237, 280, and, 302 lines of code respectively. Python, PHP, and Ruby file size are 7771, 6786, and 5703 bytes respectively. The shortest line of codes is Python because of Python scripting nature does not require closing tags on its method, function, or loop implementation as in PHP and Ruby. But when it comes to code file size, Python consumes more bytes to implement. Ruby programming style characteristic is shorter and simpler compared to the other PL, thus resulting the smallest file size on the implementation of GA. The results of program execution measurement in this research are shown in TABLE IV.

Python was used as the basis of performance measurement because it is the most widely used PL for research purposes [17]. Therefore we compare the execution time of all PL to Python. When it comes to the Web environment, as of research conducted by Jafar et al. [18], PHP outperform Python.

Based on last seed, best fitness cost, and best generation measurement results in TABLE IV we can infer that all of the tests returning the same best individual on the same generation for the same number of cities. Therefore proving that all PL executions and their flow of run are exactly the same.

During tests, Ruby proved to outperform Python and PHP in execution time of GA. Ruby's performance improvements vary from 6,5% to 9,99% while PHP performance are 0,22% to 6,09% slower over Python.

TABLE IV. MEASUREMENT RESULT

| Number of Cities | PL | Maximum (ms) | Minimum (ms) | Average (ms) | Standard Deviation | Coefficient of Variant (%) | Best Generation | Last Seed | Best Fitness Cost (Length) | Performance over Python (%) |
|---|---|---|---|---|---|---|---|---|---|---|
| 5 | PHP | 43,43 | 41,85 | 42,44 | 0,54 | 1,27 | 11 | 2516 | 30,305 | -0,22 |
| 5 | Python | 43,00 | 41,84 | 42,35 | 0,37 | 0,88 | 11 | 2516 | 30,305 | - |
| 5 | Ruby | 38,65 | 37,68 | 38,12 | 0,36 | 0,93 | 11 | 2516 | 30,305 | 9,99 |
| 6 | PHP | 38,44 | 36,00 | 37,18 | 0,84 | 2,27 | 7 | 2127 | 30,397 | -5,53 |
| 6 | Python | 35,73 | 34,94 | 35,23 | 0,26 | 0,74 | 7 | 2127 | 30,397 | - |
| 6 | Ruby | 32,95 | 31,90 | 32,55 | 0,34 | 1,04 | 7 | 2127 | 30,397 | 7,61 |
| 7 | PHP | 47,91 | 45,85 | 46,79 | 0,73 | 1,55 | 10 | 2735 | 41,135 | -1,56 |
| 7 | Python | 47,11 | 45,36 | 46,07 | 0,57 | 1,23 | 10 | 2735 | 41,135 | - |
| 7 | Ruby | 43,81 | 41,20 | 42,58 | 0,80 | 1,88 | 10 | 2735 | 41,135 | 7,57 |
| 8 | PHP | 62,23 | 58,61 | 60,14 | 1,18 | 1,97 | 9 | 3395 | 34,020 | -6,09 |
| 8 | Python | 57,25 | 56,21 | 56,69 | 0,36 | 0,63 | 9 | 3395 | 34,020 | - |
| 8 | Ruby | 53,57 | 52,10 | 53,00 | 0,52 | 0,98 | 9 | 3395 | 34,020 | 6,50 |
| 9 | PHP | 51,16 | 49,49 | 50,06 | 0,58 | 1,16 | 4 | 2891 | 49,062 | -3,56 |
| 9 | Python | 49,44 | 47,75 | 48,34 | 0,50 | 1,04 | 4 | 2891 | 49,062 | - |
| 9 | Ruby | 44,67 | 43,24 | 43,87 | 0,44 | 1,01 | 4 | 2891 | 49,062 | 9,25 |
| 10 | PHP | 72,97 | 69,72 | 71,67 | 1,15 | 1,61 | 9 | 4197 | 51,881 | -2,63 |
| 10 | Python | 71,66 | 68,56 | 69,84 | 0,85 | 1,21 | 9 | 4197 | 51,881 | - |
| 10 | Ruby | 65,19 | 62,61 | 63,99 | 0,93 | 1,46 | 9 | 4197 | 51,881 | 8,37 |

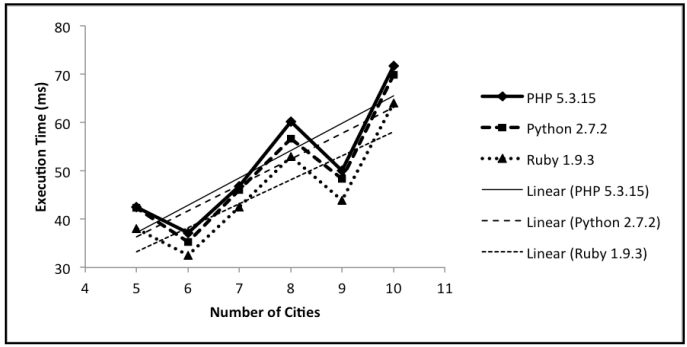

Fig. 1. Program execution time in milliseconds by number of cities

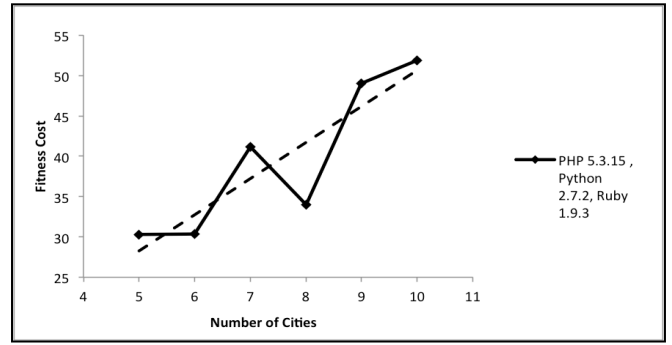

Fig. 2. Fitness cost (route's total distance) by number of cities

In TSP, expected solution of fitness cost is the minimum, the smaller the value the result will be better. Execution time grows longer when the number of cities in an individual increase as shown in Fig. 1. The relation between fitness cost (total distance on selected routes) and number of cities in an individual is shown in Fig. 2. Because of all PL were implemented in the same way, the resulting fitness cost value for the same number of cities will be the same regardless the PL used.

Regression analysis using Analysis of Variance (ANOVA) was conducted to the measurement result shown in TABLE IV. From the regression analysis result, we can infer that:

- Script execution times are highly correlated to data sizes on all implemented PL under the same GA environment. The more cities included in an individual, which mean the higher the number of inputs, the longer program execution time would be.

- It is proved that the fitness cost (route length) is highly correlated to data size on all implemented PL under the same GA environment. The higher the number of inputs (cities included in an individual), the larger the fitness cost would be. An individual which has larger fitness cost is worse.

## VI. CONCLUSION

Based on measurement and analysis process, program execution time and best fitness cost are highly relied on data size (the number of cities) in all PL. The larger the data size the longer execution time will be and the worse the outcome of GA as TSP solution.

This research GA pseudo code implementation shows that Ruby code has the smallest file size compared to Python and PHP, but Python has the least line of codes. In testing of the overall program execution time, Ruby is faster than Python and PHP. Therefore, the usage of Ruby is recommended to gain

performance in implementation of GA for TSP in a web environment over Python and PHP.


ACKNOWLEDGMENT

The authors would like to acknowledge the assistance of Program Teknologi Informasi dan Ilmu Komputer (PTIIK) and Universitas Brawijaya (UB) for their research facilities and financial support. The authors also acknowledge for the assistance of colleagues in PTIIK UB for their great assistance in improving the manuscript.